\documentclass[a4paper,fleqn]{cas-sc}
\usepackage{adjustbox}
\usepackage{float}
\usepackage[numbers]{natbib}
\usepackage[ruled,vlined]{algorithm2e}
\usepackage{graphicx}
\def\tsc#1{\csdef{#1}{\textsc{\lowercase{#1}}\xspace}}
\tsc{WGM}
\tsc{QE}
\tsc{EP}
\tsc{PMS}
\tsc{BEC}
\tsc{DE}
\begin{document}
\let\WriteBookmarks\relax
\def\floatpagepagefraction{1}
\def\textpagefraction{.001}
\shortauthors{S.Mu et~al.}

\shorttitle{Load2UnNet}    


\author[1]{Siyu Mu}[]
\ead{siyu.mu21@imperial.ac.uk}

\affiliation[1]{organization={Department of Bioengineering, Imperial College London},
                state={London},
                country={UK}}

\author[1]{Wei Xuan Chan}[]

\author[1]{Choon Hwai Yap}[orcid=0000-0003-2918-3077]
\ead{c.yap@imperial.ac.uk}

\cortext[1]{Corresponding author:}

\shorttitle{HeartUnloadNet}    

\title [mode = title]{HeartUnloadNet: A Weakly-Supervised Cycle-Consistent Graph Network for Predicting Unloaded Cardiac Geometry from Diastolic States}

\begin{abstract}
The unloaded cardiac geometry (i.e., the state of the heart devoid of luminal pressure) serves as a valuable zero-stress and zero-strain reference and is critical for personalized biomechanical modeling of cardiac function, to understand both healthy and diseased physiology and to predict the effects of cardiac interventions. However, estimating the unloaded geometry from clinical images remains a challenging task. Traditional approaches rely on inverse finite element (FE) solvers that require iterative optimization and are computationally expensive \cite{dabiri2020application}. In this work, we introduce \textit{HeartUnloadNet}, a deep learning framework that predicts the unloaded left ventricular (LV) shape directly from the end diastolic (ED) mesh while explicitly incorporating biophysical priors. The network accepts a mesh of arbitrary size along with physiological parameters such as ED pressure, myocardial stiffness scale, and fiber helix orientation, and outputs the corresponding unloaded mesh. It adopts a graph attention architecture and employs a cycle-consistency strategy to enable bidirectional (loading and unloading) prediction, allowing for partial self-supervision that improves accuracy and reduces the need for large training datasets. Trained and tested on 20,700 FE simulations across diverse LV geometries and physiological conditions, \textit{HeartUnloadNet} achieves sub-millimeter accuracy, with a Dice Similarity Coefficient (DSC) of $0.986 \pm 0.023$ and a Hausdorff Distance (HD) of $0.083 \pm 0.028$,cm, while reducing inference time to just 0.02 seconds per case, over $10^5{\times}$ faster and significantly more accurate than traditional inverse FE solvers. Ablation studies confirm the effectiveness of the architecture. Notably, the cycle-consistent design enables the model to maintain a DSC of 97\% even with as few as 200 training samples. This work thus presents a scalable and accurate surrogate for inverse FE solvers, supporting real-time clinical applications in the future. Codes are available at \url{https://github.com/SiyuMU/Loaded2UnNet}.
\end{abstract}

% Keywords

\maketitle

% Main text
\section{Introduction}
Myocardial biomechanics studies such as image-based FE modeling \cite{wang2015image, green2024myocardial} and myocardial stiffness estimation from medical images \cite{mu2025imc} are important tools that improved our understanding of cardiac physiology, and have much potential for future translation into clinical tools for disease evaluation and surgical planning.\\

In such studies, the unloaded configuration of the human LV, defined as a theoretical low stress state of the LV in the absence of intracardiac fluid pressure, serves as a fundamental reference. It underlies the computation of myocardial stress and strain, initializes FE simulations, and supports the personalization of material properties across varying loading conditions \cite{nikou2016effects,genet2016modeling}. However, this state is clinically unobservable, as the heart is continuously subjected to intracardiac pressure \cite{genet2016modeling}. Consequently, medical imaging can only capture pressure-loaded geometries, such as those observed at ED \cite{genet2016modeling}. Developing an efficient algorithm to estimate the unloaded state from cardiac geometries observable in medical images would be highly valuable for biomechanical modeling.\\

Current strategies for reconstructing the unloaded cardiac geometry rely on performing inverse FE modeling using the ED cardiac geometry reconstructed from medical images \cite{marx2022robust, wang2020efficient, krishnamurthy2013patient}. This remains the gold standard due to its strong biophysical foundations. However, it is computationally expensive, typically requiring several hours to solve a single case with a patient-specific LV mesh containing more than 1,500 nodes \cite{dabiri2020application}. Moreover, inverse FE approaches exhibit limited accuracy, particularly when confronted with poor-quality meshes or anatomical irregularities \cite{gonzales2013three}.\\

To address these limitations, the development of supervised deep learning (DL) models for estimating the unloaded cardiac state could offer significant advantages by drastically reducing computational time. Prior studies have demonstrated the feasibility of estimating the unloaded state for non-cardiac vascular structures such as the aorta \cite{liang2018machine,liang2023pytorch}, but to date, no such attempts have been reported for the heart. For clinical and research applications, it is essential that such DL models be capable of handling variations in cardiac mesh topology, myocardial helix angle configuration, tissue stiffness, and in vivo blood pressure. Furthermore, given the high cost of generating large-scale FE simulation ground truths, methods that can reduce the required dataset size while maintaining performance would be particularly beneficial.\\

Here, we introduce a graph neural network (GNN) backbone for accurately estimating the unloaded state of input cardiac meshes across varying helix angles, myocardial stiffness values, and end-diastolic pressure conditions. Unlike standard multilayer perceptrons (MLPs), which require fixed input mesh sizes and consistent node ordering, the GNN inherently encodes node connectivity and spatial relationships, allowing the model to generalize across hearts with diverse mesh resolutions and topologies \cite{kipf2016semi}. This topology-agnostic design eliminates the need for pre-processing steps such as template registration or node indexing alignment, enabling direct analysis of meshes with heterogeneous structures.\\

To reduce the required size of the FE simulation training set, we adopt a cycle-consistent training strategy for weak self-supervision, inspired by CycleGAN \cite{zhu2017unpaired}. Specifically, we enforce that the predicted unloaded geometry must accurately reconstruct the original ED shape through a reverse deformation path. This physiological reversibility provides an implicit supervisory signal, enabling the network to learn meaningful deformation mappings in a weakly supervised setting.\\

Furthermore, to fully replicate the biophysical context of inverse FE modeling, we incorporate global material parameters, such as fiber orientation and myocardial stiffness, into the network. These parameters are provided as global conditioning inputs, allowing the model to adapt its deformation predictions to patient-specific physiological conditions. By embedding such biophysical priors, our framework preserves the interpretability and material sensitivity of traditional inverse FE approaches, while retaining the computational efficiency and scalability of deep learning. These design choices collectively define \textit{HeartUnloadNet}, our proposed method for predicting unloaded cardiac geometry.\\

In summary, our contributions are fourfold:
\begin{enumerate}
    \item We developed a graph-attention network to compute the load free state of the heart from a imaged loaded state, which achieves state-of-the-art accuracy, with a mean DSC exceeding 98\% and an average HD below 0.85\,mm, outperforming previously reported both deep learning-based and inverse FE-based methods to the best of our knowledge.
    \item Our graph attention–based autoencoder naturally accommodates diverse mesh topologies across patients, enabling robust deformation modeling without requiring fixed templates or mesh alignment pre-processing.
    \item We introduce a cycle-consistent learning mechanism that enables effective training under a weakly supervised regime. Notably, using only 209 training cases, our model achieves 97\% DSC accuracy on unseen test data, highlighting its data efficiency and generalization capability.
    \item Our network is conditioned on global biophysical parameters, including fiber orientation and myocardial stiffness, and on in vivo pressures, and as such unloaded state predictions can be performed to varied combinations of these parameters.
\end{enumerate}

\section{Related Work}
\subsection{Inverse FE Methods for Unloaded Geometry Reconstruction}

Inverse FE methods constitute a fundamental and widely used approach for reconstructing unloaded cardiac geometries. Typically, the loading process from an initial estimate of the load-free state to the ED state is simulated iteratively and optimized to match the target ED pressure and cardiac shape. These FE models often represent the myocardium using anisotropic constitutive laws, such as the Fung-type formulation \cite{guccione1991passive} or the Holzapfel–Ogden model \cite{holzapfel2009constitutive}, to account for fiber directionality and tissue stiffening under diastolic loading. Iterative FE relies on updating nodal positions using optimization schemes such as Newton–Raphson solvers \cite{bathe1996finite}. However, this traditional approach frequently suffers from imperfect convergence, particularly when the initial guess deviates substantially from the true solution, and may become trapped in local minima, leading to expensive matrix updates and potential numerical instability \cite{wriggers1994non, reddy2015introduction}.\\

In recent years, many studies have aimed to enhance the efficiency and stability of inverse FE methods. Wang et al. used principal component analysis (PCA) to embed LV shapes into a low‐dimensional subspace, thereby accelerating convergence \cite{wang2020efficient}. Marx et al. proposed a relaxed, preconditioned fixed‐point iteration to jointly estimate myocardial material properties and the unloaded geometry, improving numerical stability over traditional Newton–Raphson solvers \cite{familusi2023model}. More recently, Skatulla et al. extended inverse FE techniques to a biventricular model, simultaneously inferring left and right ventricular unloaded geometries, material stiffness, and prestrain fields, all defined within a Holzapfel–Ogden framework to enhance physiological fidelity, albeit with increased computational complexity \cite{skatulla2023non}. However, despite their accuracy and interpretability, inverse FE methods remain highly resource‐intensive. The fastest approach, by Wang et al., which drastically faster than alternatives due to the use of PCA modes, still requires several minutes, whereas deep learning approaches can reduce this to the order of a second.\\

\subsection{Deep Learning Methods for Unloaded Geometry Reconstruction}
A few DL-based methods have been proposed to accelerate inverse FE for predicting unloaded geometries. Liang et al. (2018) \cite{liang2018machine} employed a basic MLP autoencoder for fully supervised prediction of unloaded vascular geometries from ED and end-systolic (ES) states. In 2023, they introduced a hybrid method that embeds a simple MLP within an inverse-FE-inspired optimization framework \cite{liang2023pytorch}. In this design, the optimizer no longer predicts full nodal displacements but instead estimates variations within a low-dimensional latent space, substantially reducing the number of parameters and achieving inference times as low as 10 seconds. While promising, this approach simplifies the vascular walls as 2D surfaces without thickness and relies on a basic MLP architecture, leaving room for performance improvement through more advanced models. To date, however, no deep learning-based method has been reported for estimating the unloaded state of the heart.

\subsection{Graph-Based Mesh Prediction for Cardiac Mechanics}
GNNs excel at modeling complex relationships on irregular meshes through graph‐structured message passing and localized feature aggregation \cite{kipf2016semi}. While direct GNN‐based approaches for unloading remain unexplored, significant progress has been made in simulating loading scenarios. Dalton et al. developed a physics‐informed GNN (PI‐GNN) emulator, an unsupervised framework that integrates graph neural networks with physics‐based constraints—to simulate Holzapfel–Ogden‐based LV meshes, achieving near–FE accuracy in predicting passive diastolic displacements \cite{dalton2022emulation,dalton2023physics}. Their model was trained on 1,000 synthetically generated LV meshes and demonstrated negligible errors in passive dilation under prescribed ED pressures. Meanwhile, Shi et al. introduced HeartSimSage \cite{shi2025heartsimsage}, a supervised GNN that incorporates attention mechanisms and Laplace–Dirichlet spatial encoding to accommodate diverse biventricular geometries, fiber orientations, and boundary conditions. Implemented on GPU, HeartSimSage achieved a 13,000× speedup over conventional FE methods, with passive displacement errors of only 0.13\% ± 0.12\%. These studies underscore the potential of GNNs, which naturally encode mesh topology and enable scalable message passing, to serve as efficient surrogates for FE simulations. However, existing methods focus on forward simulations of deformation under loading and have not yet addressed the inverse problem of reconstructing the load-free state directly from image-derived geometries.

\section{Methods}

\begin{figure}
	\centering
	\includegraphics[width=\textwidth]{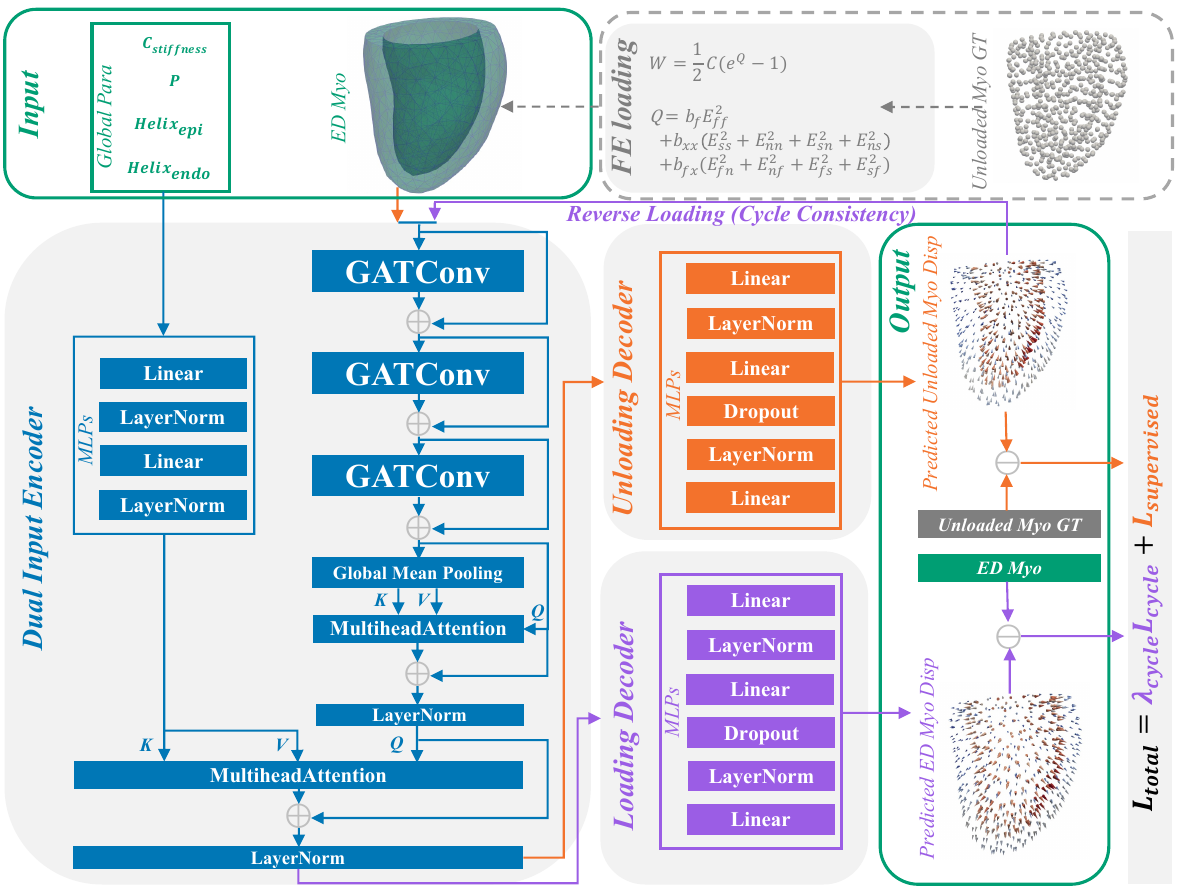}
        \caption{Schematic of the proposed \textit{HeartUnloadNet} architecture. The model takes a LV myocardial tetrahedral mesh along with physiological parameters as input. During training, a dual-input encoder extracts local features from the mesh using stacked GATConv layers and global physiological features via an MLP, which are fused through attention mechanisms. The decoder predicts both forward (ED $\rightarrow$ unloaded) and reverse (unloaded $\rightarrow$ ED) deformations, enabling a cycle-consistent training scheme. The total loss combines a supervised displacement term with a cycle consistency term.}
	\label{FIG:schematic}
\end{figure}

\subsection{Finite Element (FE) Ground Truth Generation}
Since the Loaded2UnNet framework is data-driven, we first need to establish paired ground truth for the unloaded and ED LV myocardial meshed geometries, this is achieved via a traditional FE model in which the loading model minimizes the potential energy \cite{bonet2001large}:
\begin{equation}
loss_{\text{energy}} \;=\; \int_{\Omega_{\text{myo}}} W \, dV \;-\; \int_{\Omega_{\text{endo}}} P \cdot u \, dA 
\label{equ:energy_loss}
\end{equation}

\noindent The first integral accounts for the total deformational strain energy stored in the myocardium, with $V$ denoting the myocardial volume. The second integral quantifies the work done performed by the endocardium on the blood under pressure $P$ on the LV cavity, where $u$ is the endocardial displacement from the unloaded state and $A$ is the endocardial surface area.\\

$W$ is the strain energy density. Only the passive stiffness component is considered for our task of diastolic loading, the active contractile component is ignored. The passive strain energy is computed using a Fung-type transversely isotropic hyperelastic material model that incorporates fiber orientation through the helix angle \(\theta_h\), which varies linearly from endocardium to epicardium \cite{guccione1991passive,shavik2018high}:

\begin{equation}
W \;=\; \tfrac{1}{2} C \bigl( e^Q - 1 \bigr)
\label{equ:W_passive}
\end{equation}

\begin{align}
Q &= b_f E_{ff}^2 + b_{xx} \left( E_{ss}^2 + E_{nn}^2 + E_{sn}^2 + E_{ns}^2 \right) \notag \\
&\quad + b_{fx} \left( E_{fn}^2 + E_{nf}^2 + E_{fs}^2 + E_{sf}^2 \right)
\label{equ:Q_definition}
\end{align}

\noindent where \(E_{ij}\) are the components of the Green–Lagrange strain tensor \(E\), given by \(E = \tfrac{1}{2}(F^{T}F - I)\), and the subscripts \(f\), \(s\), and \(n\) denote fiber, sheet, and sheet‐normal directions in the myocardium. The coefficients \(b_{ij}\) represent myocardial anisotropic stiffness coefficients, and \(C\) denotes the global stiffness scaling factor. The deformation gradient tensor \(F\) is defined as \(F = \frac{\partial x}{\partial X}\), with \(x\) indicating the deformed coordinates and \(X\) the unloaded coordinates.\\

We assume the helix angle varies linearly across the wall: \(\theta_h = \theta_{\mathrm{endo}} + \xi(\theta_{\mathrm{epi}} - \theta_{\mathrm{endo}})\), where \(\theta_{\mathrm{endo}}\) and \(\theta_{\mathrm{epi}}\) are the helix angles at the endocardium and epicardium, respectively, and \(\xi \in [0,1]\) is the transmural coordinate (0 at endocardium, 1 at epicardium). The unit fiber direction is \(\mathbf{a}_f(\theta_h) = [\cos\theta_h,\;\sin\theta_h,\;0]^T\). We define a scalar field \(\phi(X)\) on the myocardial mesh such that \(\phi=0\) on the endocardial surface and \(\phi=1\) on the epicardial surface, hence \(\nabla\phi\) points in the transmural (sheet‐normal) direction. The normalized transmural (sheet‐normal) direction is \(\mathbf{a}_n = \nabla\phi / \|\nabla\phi\|\), and the sheet direction is \(\mathbf{a}_s = \mathbf{a}_n \times \mathbf{a}_f\), orthogonal to both \(\mathbf{a}_f\) and \(\mathbf{a}_n\).\\ 

In the process of obtaining the ground truth, we fixed the material parameters \(b_{ij}\). The only changed global parameters parameters allowed to vary are the LV pressure \(P\), the global stiffness parameter \(C\), and the two helix angles \(\theta_{\mathrm{endo}}\) and \(\theta_{\mathrm{epi}}\). By prescribing these four variables, the FE model minimizes the potential energy using a Newton–Raphson solver: starting from the unloaded state, the Newton iterations converge to the equilibrium ED configuration under the applied LV pressure \(P\).

\subsection{HeartUnloadNet Architecture}

\begin{algorithm}
\caption{HeartUnloadNet}
\label{alg:HeartUnloadNet}
\SetKwInOut{Input}{Input}
\SetKwInOut{Output}{Output}

\Input{
Global physiological parameters $(P, C, \theta_{\mathrm{endo}}, \theta_{\mathrm{epi}})$, ED myocardial mesh $\widetilde{\mathcal{M}}_{ED} = (\mathbf{X}_e, \mathcal{E}_e)$
}

\Output{
Predicted displacement $\widehat{\Delta\mathbf{u}}_{e \to u}$
}

\BlankLine

\textbf{Ground Truth from FE Simulation: }
\BlankLine
Given global physiological parameters $(P, C, \theta_{\mathrm{endo}}, \theta_{\mathrm{epi}})$, we apply forward FE loading from $\mathcal{M}_u$ to obtain:

\Indp
\begin{tabular}{@{}ll}
$\bullet$ Ground-truth ED mesh: & $\widetilde{\mathcal{M}}_{ED} = \mathcal{M}_u + \Delta\mathbf{u}_{u \to e}$ \\
$\bullet$ Supervised pair:      & $(\mathcal{M}_u, \widetilde{\mathcal{M}}_{ED})$
\end{tabular}
\Indm
\vspace{0.5em}

\textbf{Training Process}\;

\For{$epoch = 1$ \KwTo $T$}{  
\For{$\mathbf{H}^{(0)} \in \left\{\widetilde{\mathcal{M}}_{ED},\; \widehat{\mathcal{M}}_{u} \right\}$}{
\BlankLine
\textbf{Encoder}
\BlankLine
Mesh Encoder with Global Context Injection:
\begin{equation*}
\begin{aligned}
&\mathbf{H}^{(\ell)} = \text{GAT}^{(\ell)}(\mathbf{H}^{(\ell-1)}) + \mathbf{H}^{(\ell-1)}, \quad \ell = 1, 2, 3\\
&\mathbf{g}_{\text{mesh}} = \text{MeanPool}(\mathbf{H}^{(3)}) \in \mathbb{R}^{d} \quad \text{(broadcast to } \mathbb{R}^{n \times d}) \\
&\mathbf{Z}_{\text{mesh}} = \text{LayerNorm} \left( \mathbf{H}^{(3)} + 
\big\|_{m=1}^M \text{softmax} \left( \frac{ \mathbf{Q}_m^{\text{H}}(\mathbf{K}_m^{{\text{g}_\text{mesh}}})^\top }{ \sqrt{d_m} } \right) \mathbf{V}_m^{\text{g}_\text{mesh}} \right) \in \mathbb{R}^{n \times d}\\
\end{aligned}
\end{equation*}

Global parameter encoder:
\(
\mathbf{g}_{\text{global}} = \text{MLP}^{global}([P, C, \theta_{\mathrm{endo}}, \theta_{\mathrm{epi}}])
\)

Cross-Attention Fusion: 
\[
\mathbf{Z}_{\text{Fusion}} = \text{LayerNorm} \left( \mathbf{Z}_{\text{mesh}} + 
\big\|_{m=1}^M \text{softmax} \left( \frac{ \mathbf{Q}_m^{\text{Zmesh}}(\mathbf{K}_m^{{\text{g}_\text{global}}})^\top }{ \sqrt{d_m} } \right) \mathbf{V}_m^{\text{g}_\text{global}} \right) \in \mathbb{R}^{n \times d}
\]

\textbf{Decoder}
\BlankLine
Predicted unloaded displacements ($\text{with } \mathbf{H}^{(0)} \leftarrow \widetilde{\mathcal{M}}_{\mathrm{ED}}$):
\[
\widehat{\Delta\mathbf{u}}_{e \to u} = \text{MLP}^{e \to u}(\text{Dropout}(\mathbf{Z}_{\text{Fusion}})) {\in \mathbb{R}^{n \times 3}, \quad \widehat{\mathcal{M}}_{u} = \widetilde{\mathcal{M}}_{ED} + \widehat{\Delta\mathbf{u}}_{e \to u}}
\]
\BlankLine
Predicted ED displacements ($\text{with } \mathbf{H}^{(0)} \leftarrow \widehat{\mathcal{M}}_{\mathrm{u}}$):
\[
\widehat{\Delta\mathbf{u}}_{u \to e} = \text{MLP}^{u \to e}(\text{Dropout}(\mathbf{Z}_{\text{Fusion}})) {\in \mathbb{R}^{n \times 3}, \quad \widehat{\mathcal{M}}_{ED} = \widehat{\mathcal{M}}_{u} + \widehat{\Delta\mathbf{u}}_{u \to e}}
\]
}

\textbf{Loss Functions and Parameter Update}
\[
\mathcal{L}_{\text{total}} = \left\| \widehat{\mathcal{M}}_{\mathrm{u}} - \mathcal{M}_u \right\|^2 + 
\lambda_{\text{cycle}} \cdot \left\| \widehat{\mathcal{M}}_{ED} - \widetilde{\mathcal{M}}_{ED} \right\|^2, \quad
\Theta \leftarrow \Theta - \eta \cdot \nabla_\Theta \mathcal{L}_{\text{total}}
\]

}  
\end{algorithm}

\subsubsection{Overview}

\textit{HeartUnloadNet} is a cycle-consistent graph-attention–based framework designed to infer the unloaded LV geometry from an observed ED state. As illustrated in Fig.~\ref{FIG:schematic} and Alg.~\ref{alg:HeartUnloadNet}, the architecture integrates both local geometric structures and global physiological parameters through two main components:

\begin{itemize}
    \item A \textbf{graph attention–based encoder} that integrates node-level mesh features with global physiological parameters using a structured attention mechanism. 
    \item A \textbf{cycle-consistent bi-directional decoder} that estimates nodal displacements in both directions: from the ED state to the unloaded configuration and vice versa, where the same encoder but different decoders are used for the two directions, thereby enforcing cycle consistency and structural coherence.
\end{itemize}

\subsubsection{Graph Attention-based Encoder}

The encoder comprises three main components: mesh encoder with global context injection, global parameter encoder, and cross-attention fusion.

\paragraph{\textbf{Mesh Encoder with Global Context Injection}:} 

Given an input mesh with node features $\mathbf{H}^{(0)}$, the encoder applies three sequential graph attention layers. Each graph attention layer updates node features by aggregating neighboring node information through attention weights \cite{velivckovic2017graph}:

\begin{equation}
\mathbf{H}^{(\ell)} = \text{GAT}^{(\ell)}(\mathbf{H}^{(\ell-1)}) + \mathbf{H}^{(\ell-1)} = 
\big\|_{m=1}^{M} \sigma\left( \sum_{j \in \mathcal{N}(i)} \alpha_{ij}^{(m,\ell)} \mathbf{W}^{(m,\ell)} \mathbf{H}_j^{(\ell-1)} \right)
+ \mathbf{H}^{(\ell-1)}\in \mathbb{R}^{n \times d}, \quad \ell = 1,2,3
\end{equation}

\begin{equation}
\alpha_{ij}^{(m,\ell)} = 
\frac{
\exp\left( \text{LeakyReLU} \left( {\mathbf{a}^{(m,\ell)}}^\top \left[ \mathbf{W}^{(m,\ell)}\mathbf{H}_i^{(\ell-1)} \,\|\, \mathbf{W}^{(m,\ell)}\mathbf{H}_j^{(\ell-1)} \right] \right) \right)
}{
\sum\limits_{k \in \mathcal{N}(i)} 
\exp\left( \text{LeakyReLU} \left( {\mathbf{a}^{(m,\ell)}}^\top \left[ \mathbf{W}^{(m,\ell)}\mathbf{H}_i^{(\ell-1)} \,\|\, \mathbf{W}^{(m,\ell)}\mathbf{H}_k^{(\ell-1)} \right] \right) \right)
}
\label{eq:gat_attention}
\end{equation}

\noindent where \( \mathbf{H}^{(\ell)} \in \mathbb{R}^{n \times d} \) is the node feature matrix at the \( \ell \)-th GAT layer, with each row \( \mathbf{H}_i^{(\ell)} \) representing node \( i \). Here, \( n \) denotes the number of mesh nodes, and \( d \) is the dimensionality of node features. In our setting, \( \mathbf{H}^{(0)} \) is initialized using the normalized 3D coordinates from the input ED mesh \( \widetilde{\mathcal{M}}_{\mathrm{ED}} \). The operator \( \text{GAT}^{(\ell)} \) refers to a multi-head Graph Attention Network layer that aggregates information from neighboring nodes via learned attention weights. The concatenation operator \( \big\|_{m=1}^{M} \) denotes the output combination across \( M \) independent attention heads, each followed by a nonlinear activation function \( \sigma(\cdot) \), instantiated as ReLU in our case. The set \( \mathcal{N}(i) \) denotes the 1-hop neighbors of node \( i \), implicitly encoding the mesh connectivity. The term \( \alpha_{ij}^{(m,\ell)} \) represents the normalized attention coefficient associated with edge \( (i,j) \) in the \( m \)-th attention head at layer \( \ell \), reflecting the relative importance of node \( j \) to node \( i \). These coefficients are computed using a softmax operation over all neighbors \( j \in \mathcal{N}(i) \), as described in Eqn~\ref{eq:gat_attention}. The attention parameters \( \mathbf{a}^{(m,\ell)} \) and transformation matrices \( \mathbf{W}^{(m,\ell)} \) are learnable for each attention head and layer. The addition of \( \mathbf{H}^{(\ell-1)} \) serves as a residual connection, enabling the network to preserve low-level features and improving training stability by mitigating vanishing gradients.\\

After the final GAT layer, node-level features $\mathbf{H}^{(3)}$ are aggregated into a global mesh-level representation $\mathbf{g}_{\text{mesh}}$ via mean pooling \cite{lecun1989backpropagation}. This global vector encodes the overall geometry and feature distribution of the entire mesh, providing essential context for subsequent decoding. By incorporating this global information, the decoder is expected to better capture whole-organ structural properties and implicitly enforce FE–like constraints, such as volume preservation and global energy consistency.
\begin{equation}
\mathbf{g}_{\text{mesh}} = \frac{1}{n} \sum_{i=1}^{n} \mathbf{H}_i^{(3)} 
\in \mathbb{R}^{d}
\end{equation}

\noindent where $n$ is the number of nodes in the mesh, and $\mathbf{g}_{\text{mesh}}$ is then broadcast back to each node to match the node-level dimension for subsequent fusion operations.\\

To incorporate global context into each node's local representation, we employ a cross-attention mechanism \cite{vaswani2017attention}, where the mesh node features $\mathbf{H}^{(3)}$ serve as queries $\mathbf{Q}_m^{\mathbf{H}}$, and the broadcasted global vector $\mathbf{g}_{\text{mesh}}$ provides keys $\mathbf{K}_m^{\mathbf{g}_{\text{mesh}}}$ and values $\mathbf{V}_m^{\mathbf{g}_{\text{mesh}}}$. Specifically, the fused representation is computed as:

\begin{equation}
\mathbf{Z}_{\text{mesh}} = \text{LayerNorm} \left( \mathbf{H}^{(3)} + \big\|_{m=1}^{M} \text{softmax} \left( \frac{ \mathbf{Q}_m^{\mathbf{H}} (\mathbf{K}_m^{\mathbf{g}_{\text{mesh}}})^\top }{ \sqrt{d_m} } \right) \mathbf{V}_m^{\mathbf{g}_{\text{mesh}}} \right) \in \mathbb{R}^{n \times d}
\end{equation}

\begin{equation}
\begin{aligned}
\mathbf{Q}_m^{\mathbf{H}} &= \mathbf{H}^{(3)} \mathbf{W}_Q^{(m)} \in \mathbb{R}^{n \times d_m}, \quad \mathbf{K}_m^{\mathbf{g}_{\text{mesh}}}= \mathbf{g}_{\text{mesh}} \mathbf{W}_K^{(m)} \in \mathbb{R}^{1 \times d_m}, \quad \mathbf{V}_m^{\mathbf{g}_{\text{mesh}}}=\mathbf{g}_{\text{mesh}} \mathbf{W}_V^{(m)} \in \mathbb{R}^{1 \times d_m}
\end{aligned}
\end{equation}

\noindent where \( \mathbf{H}^{(3)} \in \mathbb{R}^{n \times d} \) is the node feature matrix after the third GAT layer, where each row \( \mathbf{H}_i^{(3)} \) corresponds to node \( i \). The operator \( \text{LayerNorm}(\cdot) \) denotes layer normalization applied across the feature dimension, which normalizes each node’s feature vector independently to stabilize training and improve convergence \cite{ba2016layer}. The matrices \( \mathbf{W}_Q^{(m)}, \mathbf{W}_K^{(m)}, \mathbf{W}_V^{(m)} \in \mathbb{R}^{d \times d_m} \) are learnable projection matrices for the \( m \)-th attention head, where \( d_m \) is the feature dimensionality of each head. \\

This step allows each node to selectively attend to different aspects of the global mesh embedding, enabling feature modulation conditioned on the overall geometry. Such cross-attention enhances the network’s capacity to model dependencies beyond local neighborhoods, crucial for biomechanical systems,  where global shape and boundary-aware context jointly influence tissue behavior.

\paragraph{\textbf{Global Parameter Encoder}:} 

To incorporate global physiological priors into the model, we encode the pressure–stiffness–orientation tuple \((P, C, \theta_{\mathrm{endo}}, \theta_{\mathrm{epi}})\) into a compact feature vector via a MLP):

\[
\mathbf{g}_{\text{global}} = \text{MLP}^{\text{global}} \left( [P,\ C,\ \theta_{\mathrm{endo}},\ \theta_{\mathrm{epi}}] \right) \in \mathbb{R}^{d}
\]

It is noting that each of the four physiological parameters is normalized to the range \([0, 1]\) based on its respective physical bounds prior to encoding, with detailed values provided in the following section. This normalization promotes stable training and ensures balanced contributions from each parameter.

\paragraph{\textbf{Cross-Attention Fusion}:} 

The mesh-level features $\mathbf{Z}_{\text{mesh}}$ and global physiological parameters $(P, C, \theta_{\mathrm{endo}}, \theta_{\mathrm{epi}})$ are integrated via a cross-attention fusion:

\begin{equation}
\mathbf{Z}_{\text{Fusion}} = \text{LayerNorm}\left(\mathbf{Z}_{\text{mesh}} + \big\|_{m=1}^M \text{softmax}\left(\frac{\mathbf{Q}_m^{Z{\text{mesh}}}(\mathbf{K}_m^{g_{\text{global}}})^\top}{\sqrt{d_m}}\right)\mathbf{V}_m^{g_{\text{global}}}\right)
\end{equation}

\noindent where $\mathbf{Q}_m^{Z_{\text{mesh}}}$, $\mathbf{K}_m^{g_{\text{global}}}$, and $\mathbf{V}_m^{g_{\text{global}}}$ are the query, key, and value matrices for the $m$-th attention head, computed from $\mathbf{Z}_{\text{mesh}}$ and $\mathbf{g}_{\text{global}}$, respectively. The cross-attention mechanism enables each node to selectively attend to the global physiological context.

\subsubsection{Cycle-Consistent Bi-Directional Decoder}

To achieve optimal performance while minimizing the required number of ground truth samples, we propose a cycle-consistent bi-directional decoder that estimates nodal displacements in both directions: ED to unloaded $(e \to u)$ and unloaded to ED $(u \to e)$. This is inspired by the principle of CycleGAN\cite{zhu2017unpaired}, where an image-to-image translation network is trained bidirectionally, with cycle consistency constraints imposed to ensure that an image can be faithfully reconstructed after a forward and subsequent backward translation. This cycle consistency provides a form of weak self-supervision that regularizes learning, enhances generalization, and mitigates overfitting, thereby improving performance and reducing the required training sample size. Here, we introduce a similar reconstruction loop that enables self-supervised learning: the network predicts the unloaded mesh from the ED state, then reconstructs the ED mesh from the predicted unloaded state. Minimizing the discrepancy between the cyclic reconstructed and original ED mesh provides this additional self-supervision.\\

Formally, a shared encoder is employed for both directions, ED to unloaded and unloaded back to ED, to extract a unified fused feature representation \( \mathbf{Z}_{\text{Fusion}} \). Two distinct MLP decoders are subsequently used to predict displacements in each direction from \( \mathbf{Z}_{\text{Fusion}} \):
\begin{equation}
\widehat{\Delta\mathbf{u}}_{e \to u} = \text{MLP}^{e \to u}(\text{Dropout}(\mathbf{Z}_{\text{Fusion}})) \in \mathbb{R}^{n \times 3}, \quad
\widehat{\mathcal{M}}_{u} = \widetilde{\mathcal{M}}_{\mathrm{ED}} + \widehat{\Delta\mathbf{u}}_{e \to u}
\end{equation}

\begin{equation}
\widehat{\Delta\mathbf{u}}_{u \to e} = \text{MLP}^{u \to e}(\text{Dropout}(\mathbf{Z}_{\text{Fusion}})) \in \mathbb{R}^{n \times 3}, \quad
\widehat{\mathcal{M}}_{\mathrm{ED}} = \widehat{\mathcal{M}}_{u} + \widehat{\Delta\mathbf{u}}_{u \to e}
\end{equation}

\noindent where $\widehat{\Delta\mathbf{u}}_{e \to u}, \widehat{\Delta\mathbf{u}}_{u \to e} \in \mathbb{R}^{n \times 3}$ are the predicted displacement vectors at each node. $\widetilde{\mathcal{M}}_{\mathrm{ED}}$ denotes the input ED mesh geometry, and $\widehat{\mathcal{M}}_{u}, \widehat{\mathcal{M}}_{\mathrm{ED}}$ are the predicted unloaded and reconstructed ED meshes, respectively. $\text{Dropout}(\cdot)$ applies dropout regularization to the fused features during training, helping to prevent overfitting by randomly deactivating a fraction of the feature dimensions \cite{srivastava2014dropout}.

\subsubsection{Loss Function}

Based on the predicted meshes in both directions, we define the total loss that combines supervised displacement regression with cycle consistency as:
\[
\mathcal{L}_{\text{total}} = \left\| \widehat{\mathcal{M}}_{\mathrm{u}} - \mathcal{M}_u \right\|^2 + 
\lambda_{\text{cycle}} \cdot \left\| \widehat{\mathcal{M}}_{ED} - \widetilde{\mathcal{M}}_{ED} \right\|^2
\]

\noindent where $\lambda_{\text{cycle}}$ controls the strength of the cycle-consistency constraint.

\subsection{Experimental Setup}
\subsubsection{Data Acquisition}
We constructed a dataset of LV geometries based on a statistical shape model derived from principal component analysis (PCA) of the \textit{Cardiac Atlas Project (CAP)} (\url{https://www.cardiacatlas.org/left-ventricular-modes/}). The model was trained on 1,991 asymptomatic subjects from the MESA cohort \cite{medrano2014left} and captures dominant modes of anatomical variation at ED, including size, sphericity, and concentricity. To synthesize anatomically plausible ED geometries, we sampled PCA weights from a standard normal distribution and truncated each mode to \([-2, +2]\), corresponding to approximately 95\% of population-level variation due to the zero mean and unit variance of each mode. We randomly generated 60 distinct ED shapes by drawing unique coefficient sets from this distribution. The resulting surface meshes were then converted into volumetric tetrahedral meshes using Gmsh \cite{geuzaine2009gmsh}. Due to meshing variability, the final FE meshes did not share an identical node structure, node counts ranged from 600 to 1000, and tetrahedral elements from 1800 to 2500. All meshes were visually inspected to ensure anatomical plausibility and sufficient geometric quality for downstream simulation.\\

% \begin{figure}
% 	\centering
% 	\includegraphics[width=\textwidth]{figs/supervised_pair.pdf}
%         \caption{Generation of supervised training pairs via FE-based unloading–reloading simulation. The resulting pair \((\mathcal{M}_u, \widetilde{\mathcal{M}}_{ED})\) forms a physically consistent supervision target.}
% 	\label{FIG:supervised_pair}
% \end{figure}

After mesh generation, we constructed supervised training pairs \((\mathcal{M}_u, \widetilde{\mathcal{M}}_{ED})\) using a FE unloading–reloading strategy. The unloaded geometry is first obtained through inverse FE, and then the loading is reapplied to the unloaded geometries to obtain the ED geometry. The unloaded and ED pair from the reloading exercise is used for training, while the original ED geometry is discarded. This is because the inverse FE unloading process is typically under-constrained and can have non-unique solution, while the forward FE loading process deterministically generates the ED geometry. The resulting supervision pair is thus physically consistent and fully defined by the prescribed loading parameters.\\

\begin{table}[htbp]
\centering
\caption{Simulation parameters used in the FE pipeline. Pressure, fiber orientation, and the Fung stiffness scaling coefficient \(C\) were systematically varied, while the remaining dimensionless stiffness ratios \(b_{ff}, b_{xx}, b_{fx}\) were fixed across all simulations.}
\label{TAB:parameters}
\begin{tabular}{ll}
\toprule
\textbf{Parameter} & \textbf{Values} \\
\midrule
\multicolumn{2}{l}{\textit{Varied across simulations}} \\
\hspace{1em}End-diastolic pressure (EDP) & \{4, 6, 8, 10, 12, 14\} mmHg \\
\hspace{1em}Fung stiffness scaling coefficient \(C\) & \{50, 100, 150, 200, 250, 300\} Pa \\
\hspace{1em}Endocardial fiber angle \(\theta_{\mathrm{endo}}\) & \{60°, 65°, 70°\} \\
\hspace{1em}Epicardial fiber angle \(\theta_{\mathrm{epi}}\) & \{-60°, -65°, -70°\} \\
\midrule
\multicolumn{2}{l}{\textit{Fixed across simulations}} \\
\hspace{1em}Fung anisotropic coefficients & \(b_{ff}=29.9,\ b_{xx}=13.3,\ b_{fx}=26.6\) \\
\bottomrule
\end{tabular}
\end{table}

To simulate diverse physiological scenarios, we systematically varied key input parameters as detailed in Table~\ref{TAB:parameters} during the unloading–reloading process. This combinatorial design resulted in a total of 20,700 simulation cases, spanning both anatomical shape and physiological conditions. To evaluate model generalization, we designed two training–testing splits targeting distinct sources of variability.

\paragraph{\textbf{Shape Splits}:} 
To evaluate the model’s ability to unseen anatomical geometries, we split the 60 ED shapes into 42 for training (13,890 samples, 67\%) and 18 for testing (6,810 samples, 33\%). All global parameter combinations associated with each shape were preserved within the same split to prevent shape leakage. This ensured that test-time geometries were entirely unseen during training, while the model still operated within familiar physiological ranges. This protocol reflected a typical deployment scenario in clinical applications, where patient anatomies were novel but physiological parameters lay within the trained distributions. This shape-based split served as the default evaluation setting and was consistently used across all ablation and comparison experiments unless otherwise noted.

\paragraph{\textbf{Global Parameter Splits}:} 
To evaluate the model’s robustness to unseen physiological conditions, we adopted a leave-one-value-out (LOVO) cross-validation strategy for each global parameter \((P, C, \theta_{\mathrm{endo}}, \theta_{\mathrm{epi}})\). In each run, one specific parameter value was held out during training and used exclusively for testing, while all other values were retained. This setup allowed a controlled assessment of the model’s ability to generalize across untrained physiological conditions. All parameter values were normalized based on the training set prior to model input.

\subsection{Implementation Details}

\paragraph{\textbf{Software and hardware}:}
All experiments were implemented in Python using PyTorch and PyTorch Geometric. Training and inference were conducted on a workstation equipped with an NVIDIA RTX 3090 GPU (24\,GB VRAM). CUDA (v11.1) and cuDNN optimizations were enabled.

\paragraph{\textbf{Training hyperparameters}:} 
The model was trained using the AdamW optimizer with a learning rate of 0.001 and weight decay of $10^{-4}$. A total of 1000 training epochs were used with early stopping (patience = 50). Dropout was set to 0.1 in all attention and decoder layers. Mixed-precision training was disabled to ensure numerical stability. The hidden dimension was set to 128, with 4 attention heads used throughout the network. The cycle consistency loss was weighted by 0.2. Global parameters $(P, C, \theta_{\mathrm{endo}}, \theta_{\mathrm{epi}})$ were normalized to $[0,1]$ before being input into the encoder, and all mesh coordinates were globally normalized to have zero mean and unit scale.

\paragraph{\textbf{Evaluation metrics}:}
Model performance was evaluated using four geometric metrics between predicted and ground-truth unloaded meshes, as summarized in Table~\ref{tab:evaluation-metrics}.  It should be noted that the DSC was adapted to a node-level formulation, computed as the percentage of mesh nodes whose Euclidean distance error fell below a 0.1 mm threshold. This leverages one-to-one correspondence and identical topology to assess overlap directly in mesh space without voxelization, thus preserving spatial resolution and enhancing sensitivity to subtle differences. Such node-level, thresholded overlap metrics have been previously adopted in mesh-based simulations such as~\cite{whitaker2019improved}.

\begin{table}[htbp]
\centering
\caption{Evaluation metrics used for geometric and topological comparison.}
\label{tab:evaluation-metrics}
\begin{tabular}{@{}ll@{}}
\toprule
\textbf{Metric} & \textbf{Description} \\
\midrule
Dice Similarity Coefficient (DSC) & Proportion of nodes with error $<$\,0.1\,mm based on one-to-one correspondence \\
Hausdorff distance (HD) & Maximum point-wise deviation between predicted and ground-truth surfaces \\
Mean distance (MD) & Mean Euclidean distance between corresponding nodes (global accuracy) \\
Standard deviation (SD) & Standard deviation of node-wise distance errors (spatial consistency) \\
\bottomrule
\end{tabular}
\end{table}

\section{Results}
\subsection{Quantitative Performance Comparison}

\begin{table}[htbp]
\centering
\caption{
Quantitative performance comparison of baseline methods for predicting unloaded LV geometry from the ED state. 
}
\label{tab:baseline-comparison}
\begin{tabular}{@{}lccccc@{}}
\toprule
\textbf{Method} & \textbf{DSC ↑} & \textbf{HD ↓ (cm)} & \textbf{MD ↓ (cm)} & \textbf{SD ↓ (cm)} & \textbf{Time ↓ (s)} \\
\midrule
GCN & 0.674$\pm$0.169 & 0.263$\pm$0.073 & 0.070$\pm$0.017 & 0.041$\pm$0.010 & 0.021$\pm$0.002 \\
PointNet++ & 0.602$\pm$0.312 & 0.216$\pm$0.085 & 0.093$\pm$0.048 & 0.038$\pm$0.017 & \textbf{0.018$\pm$0.002} \\
Inverse FE & 0.811$\pm$0.094 & 0.185$\pm$0.080 & 0.097$\pm$0.022 & 0.099$\pm$0.010 & 7238.1$\pm$678.2 \\
\textbf{\textit{HeartUnloadNet} (ours)} & \textbf{0.986$\pm$0.023} & \textbf{0.083$\pm$0.028} & \textbf{0.028$\pm$0.010} & \textbf{0.013$\pm$0.004} & 0.023$\pm$0.002 \\
\bottomrule
\end{tabular}
\end{table}

Table~\ref{tab:baseline-comparison} summarized the quantitative performance of our proposed \textit{HeartUnloadNet} against three baseline methods: a graph convolutional network (GCN)\cite{kipf2016semi}, PointNet++\cite{qi2017pointnetplusplus}, and a conventional inverse FE solver\footnote{\url{https://github.com/WeiXuanChan/heartFEM}} that applied the widely used backward displacement method by Finsberg et al.~\cite{finsberg2018efficient}, which iteratively estimated unloaded geometry by simulating inverse FE unloading and then forward reloading to match target pressure conditions. \textit{HeartUnloadNet} consistently outperformed all baselines across all four geometric metrics, achieving the highest DSC ($0.986 \pm 0.023$) and the lowest HD ($0.083 \pm 0.028$), MD ($0.028 \pm 0.010$), and SD ($0.013 \pm 0.004$).\\

Notably, GCN outperformed PointNet++ (e.g., DSC $0.674$ vs.\ $0.602$), likely due to its use of mesh connectivity, which implicitly encoded structural information such as neighborhood topology and element volume, important for preserving biomechanical consistency. Building upon this advantage, we adopted the GCN framework as the architectural backbone of \textit{HeartUnloadNet}, progressively integrating pooling, co-attention fusion, and cycle consistency modules to enhance both geometric fidelity and physiological awareness.\\

Furthermore, compared to the conventional inverse FE solver, which took over 7,000 seconds per case or roughly 2 hours, \textit{HeartUnloadNet} achieved markedly better geometric accuracy (DSC: \textbf{0.986} vs.\ 0.811) while reducing inference time by more than five orders of magnitude ($\sim$0.02s). Notably, the inverse FE method exhibited limited accuracy, with a DSC of 0.81 and HD of 0.185. This was likely due to its first-order approximations when generating unloading deformations, designed to reduce computational burden and improve convergence, but the technique nonetheless required substantial time. As such, despite the FE method’s strong physical grounding, it had limited practicality for large-scale or real-time applications. In contrast, \textit{HeartUnloadNet} offered near-instantaneous prediction with FE-level fidelity, making it far more suitable for clinical deployment or simulation-intensive tasks.

\subsection{Weakly Supervised Learning Performance}
To investigate the capability of the cycle consistency in \textit{HeartUnloadNet} to enable weak or reduced supervision,, we conducted a controlled experiment where the supervision ratio (SR) (i.e., the percentage of labeled training samples) was progressively reduced from full supervision (100\%) down to an extremely low partial supervision (1\%). Table~\ref{tab:weak-supervision} reported the model's unloaded-state geometric prediction accuracy under each setting. We compared the fully supervised model (A0), which included the cycle consistency loss, against a variant (A1) that disabled this loss component. Both models were evaluated under identical SR conditions.\\

As shown in the table, \textit{HeartUnloadNet} (A0) maintained robust performance even under significantly reduced supervision. With just 3\% labeled data (around 200 cases), A0 still achieved a DSC of $0.971 \pm 0.045$, demonstrating strong generalization with minimal supervision. In contrast, A1, which lacked the cycle loss, suffered a drastic performance decline under the same condition (DSC dropped to $0.712 \pm 0.210$, and HD increased from $0.080$ to $0.174$). This highlighted the critical role of cycle consistency as a form of implicit regularization that enforced geometric coherence and helped prevent overfitting when direct supervision was scarce. These findings suggested that cycle consistency was not merely auxiliary, but essential for learning stable and accurate mappings in weakly supervised deformation modeling.

\begin{table}[htbp]
\centering
\caption{Performance under varying levels of supervision. We compared the full model \textit{HeartUnloadNet} (A0) with a variant (A1) that removes the cycle consistency loss. Each model was trained under decreasing SR, from full (100\%) to extreme low-supervision (1\%).}
\label{tab:weak-supervision}
\begin{tabular}{@{}llcccc@{}}
\toprule
\textbf{Model ID} & \textbf{Supervision Ratio (SR)} & \textbf{DSC ↑} & \textbf{HD ↓ (cm)} & \textbf{MD ↓ (cm)} & \textbf{SD ↓ (cm)} \\
\midrule
\multirow{5}{*}{A0: \textit{HeartUnloadNet}} 
& 100\% & $\mathbf{0.986 \pm 0.023}$ & $\mathbf{0.083 \pm 0.028}$ & $\mathbf{0.028 \pm 0.010}$ & $\mathbf{0.013 \pm 0.004}$ \\
& 30\%  & $0.982 \pm 0.028$ & $0.089 \pm 0.031$ & $0.029 \pm 0.011$ & $0.014 \pm 0.005$ \\
& 15\%  & $0.977 \pm 0.038$ & $0.081 \pm 0.030$ & $0.030 \pm 0.011$ & $0.013 \pm 0.005$ \\
& 3\%   & $0.971 \pm 0.045$ & $0.085 \pm 0.031$ & $0.032 \pm 0.011$ & $0.014 \pm 0.005$ \\
& 1\%   & $0.619 \pm 0.251$ & $0.188 \pm 0.045$ & $0.070 \pm 0.024$ & $0.033 \pm 0.008$ \\
\midrule
\multirow{5}{*}{A1: A0 w/o cycle loss} 
& 100\% & $\mathbf{0.933 \pm 0.143}$ & $\mathbf{0.080 \pm 0.037}$ & $\mathbf{0.033 \pm 0.017}$ & $\mathbf{0.014 \pm 0.007}$ \\
& 30\%  & $0.906 \pm 0.126$ & $0.101 \pm 0.037$ & $0.039 \pm 0.015$ & $0.017 \pm 0.008$ \\
& 15\%  & $0.888 \pm 0.118$ & $0.133 \pm 0.032$ & $0.043 \pm 0.013$ & $0.021 \pm 0.006$ \\
& 3\%   & $0.712 \pm 0.210$ & $0.174 \pm 0.048$ & $0.062 \pm 0.019$ & $0.030 \pm 0.008$ \\
& 1\%   & $0.514 \pm 0.197$ & $0.228 \pm 0.068$ & $0.084 \pm 0.023$ & $0.046 \pm 0.015$ \\
\bottomrule
\end{tabular}
\end{table}

\subsection{Ablation Studies}
\begin{table}[htbp]
\centering
\caption{Stepwise ablation study of key architectural components in \textit{HeartUnloadNet}. Models A1–A5 were derived by progressively removing individual modules from the full model A0, highlighting the contribution of each component to overall performance.}
\label{tab:ablation-study}
\begin{tabular}{@{}llcccc@{}}
\toprule
\textbf{Model ID} & \textbf{Components Included} & \textbf{DSC ↑} & \textbf{HD ↓ (cm)} & \textbf{MD ↓ (cm)} & \textbf{SD ↓ (cm)} \\
\midrule
A0 & Full \textit{HeartUnloadNet}                    & $\mathbf{0.986 \pm 0.023}$ & $0.083 \pm 0.028$ & $\mathbf{0.028 \pm 0.010}$ & $\mathbf{0.013 \pm 0.004}$ \\
A1 & A0 – Cycle Consistency                     & $0.933 \pm 0.143$ & $\mathbf{0.080 \pm 0.037}$ & $0.033 \pm 0.017$ & $0.014 \pm 0.007$ \\
A2 & A1 – Co-attention Fusion                   & $0.930 \pm 0.106$ & $0.094 \pm 0.024$ & $0.045 \pm 0.009$ & $0.015 \pm 0.005$ \\
A3 & A2 – Average Pooling                       & $0.823 \pm 0.180$ & $0.135 \pm 0.054$ & $0.049 \pm 0.017$ & $0.024 \pm 0.008$ \\
A4 & A3 – Graph Attention (GAE)                 & $0.778 \pm 0.165$ & $0.163 \pm 0.035$ & $0.053 \pm 0.015$ & $0.027 \pm 0.006$ \\
A5 & A4 – Dual Input (GCN base model only)      & $0.634 \pm 0.169$ & $0.263 \pm 0.073$ & $0.070 \pm 0.017$ & $0.041 \pm 0.010$ \\
\bottomrule
\end{tabular}
\end{table}

\begin{table}[htbp]
\centering
\caption{Component-wise Ablation Study of key architectural components in \textit{HeartUnloadNet}. Models C0–C6 replaced encoder, decoder, or pooling modules.}
\label{tab:component-replacement}
\resizebox{\textwidth}{!}{%
\begin{tabular}{@{}lllcccc@{}}
\toprule
\textbf{Model ID} & \textbf{Replaced} & \textbf{Alternative} & \textbf{DSC ↑} & \textbf{HD ↓ (cm)} & \textbf{MD ↓ (cm)} & \textbf{SD ↓ (cm)} \\
\midrule
A0 & – & – & $\mathbf{0.986 \pm 0.023}$ & $0.083 \pm 0.028$ & $0.028 \pm 0.010$ & $\mathbf{0.013 \pm 0.004}$ \\
C0 & GAT Encoder & GATv2 Encoder & $0.985 \pm 0.022$ & $\mathbf{0.074 \pm 0.023}$ & $\mathbf{0.028 \pm 0.009}$ & $0.013 \pm 0.005$ \\
C1 & GAT Encoder & GATv2 + 3\% SR & $0.600 \pm 0.246$ & $0.177 \pm 0.043$ & $0.074 \pm 0.025$ & $0.034 \pm 0.009$ \\
C2 & GAT Encoder & GraphSAGE Encoder & $0.963 \pm 0.056$ & $0.084 \pm 0.030$ & $0.032 \pm 0.012$ & $0.015 \pm 0.006$ \\
C3 & GAT + Pooling & MeshGraphNet & $0.942 \pm 0.029$ & $0.088 \pm 0.029$ & $0.030 \pm 0.011$ & $0.017 \pm 0.005$ \\
C4 & GAT Encoder & Graph Transformer & $0.911 \pm 0.124$ & $0.095 \pm 0.029$ & $0.039 \pm 0.015$ & $0.018 \pm 0.006$ \\
C5 & MLP Decoder & GAT Decoder & $0.725 \pm 0.204$ & $0.226 \pm 0.058$ & $0.060 \pm 0.020$ & $0.034 \pm 0.008$ \\
C6 & Average Pooling & Max Pooling & $0.934 \pm 0.111$ & $0.089 \pm 0.028$ & $0.035 \pm 0.014$ & $0.016 \pm 0.006$ \\
\bottomrule
\end{tabular}
}
\end{table}

To demonstrate the optimization of \textit{HeartUnloadNet}’s design, we performed ablation studies in this subsection. We first conducted a stepwise ablation starting from the full model (A0), progressively removing or altering key modules in variants A1–A5, as shown in Table~\ref{tab:ablation-study}. Next, we carried out a component replacement study, where each of the variants C0–C6 had a key module (encoder, decoder, or pooling) replaced (Table~\ref{tab:component-replacement}).\\

In the progressive ablation study (Table~\ref{tab:ablation-study}), the cyclic consistency loss was removed from A1, resulting in a clear performance degradation (DSC dropped from $0.986$ to $0.933$) and a pronounced increase in variability ($\pm$0.143), suggesting its critical role in enforcing geometric coherence and stabilizing training. HD showed a slight but negligible improvement. In A2, replacing the co-attention fusion with naive feature concatenation deteriorated HD and MD with minimal effects on DSC, underscoring the necessity of structured global–local feature alignment for accurate deformation prediction.\\

In A3, removing the pooling mechanism eliminated global mesh context, leading to a substantial reduction in DSC and an increase in HD, which highlighted the importance of hierarchical feature aggregation. In A4, replacing the GAT encoder with a standard GCN resulted in a further substantial decline in performance across all metrics, underscoring the superiority of attention-based neighborhood modeling in capturing important local features. Finally, A5 removed the dual-input structure and simply broadcast global physiological parameters to each node. This variant yielded the poorest performance (DSC: $0.634 \pm 0.169$), underlining the necessity of disentangled and context-aware fusion between mesh geometry and physiological priors for accurate biomechanical inference.\\

In the component replacement study (Table~\ref{tab:component-replacement}), C0 replaced the original GAT encoder with GATv2~\cite{brody2021attentive}, and consequently achieved slightly reduced DSC accuracy (DSC: $0.985$ vs.\ $0.986$) and slightly better HD ($0.074$ vs.\ $0.083$). This suggested that GAT and GATv2 performed similarly when sufficient supervision was available. However, in C1, using the same GATv2 under only 3\% supervision led to a major drop in performance (DSC: $0.600 \pm 0.246$). This was likely because GATv2 computed attention weights using both source and target node features through a learnable function, making it more flexible but also more prone to overfitting when data was scarce. In contrast, the original GAT used a shared attention mechanism that offered stronger inductive bias and greater stability under weak supervision. These findings reaffirmed that high-capacity architectures like GATv2 were less suitable in low-data regimes, as their flexibility increased the risk of overfitting despite regularization.\\

Additionally, replacing the encoder with GraphSAGE~\cite{hamilton2017inductive} (C2) or MeshGraphNet~\cite{pfaff2020learning} (C3) led to a noticeable decline in overall performance. These methods, while effective in other graph domains, lacked the fine-grained attention mechanism of GAT that may have been important for capturing irregular mesh geometry and directional anisotropy in biomechanical structures. Similarly, substituting the encoder with a graph transformer~\cite{yun2019graph} (C4) resulted in further degradation (DSC: $0.911$), likely due to the absence of localized inductive bias and the model’s reliance on large datasets for effective generalization. In contrast, GAT’s edge-aware attention enabled adaptive weighting of neighboring nodes based on geometric context, making it particularly well-suited for spatially structured data such as tetrahedral heart meshes.\\

Moreover, in C5, replacing the original MLP decoder with a GAT-based decoder led to substantial degradation across all metrics (e.g., SD increased from $0.013$ to $0.034$, DSC dropped to $0.725$). This suggested that additional message passing during decoding was not only unnecessary but potentially detrimental. Since the decoder operated on already fused, spatially structured features, further aggregation likely introduced redundant interactions and amplified local noise. The higher parameter complexity of GAT may also have destabilized training, especially under biomechanical constraints that favored smooth, globally coherent deformations. In C6, substituting average pooling with max pooling resulted in moderate degradation, indicating that average pooling better preserved subtle global geometry.\\

These findings highlighted the optimality of the proposed \textit{HeartUnloadNet} configurations.

\begin{table}[htbp]
\centering
\caption{LOVO cross-validation results. Each variable group (pressure, stiffness, fiber orientation) is excluded during training to evaluate generalization.}
\label{tab:lovo_results}
\begin{tabular}{@{}llccc@{}}
\toprule
\textbf{Parameter Group} & \textbf{Test Case} & \textbf{DSC ↑} & \textbf{HD ↓ (cm)} & \textbf{MD ↓ (cm)} \\
\midrule
\multirow{3}{*}{Pressure (\(P\))} 
& P (4 mmHg)   & $0.999$ & $0.007$ & $0.003$ \\
& P (6 mmHg)   & $0.999$ & $0.008$ & $0.003$ \\
& P (10 mmHg)  & $0.999$ & $0.007$ & $0.002$ \\
\midrule
\multirow{3}{*}{Stiffness (\(C\))} 
& C (50)       & $0.942$ & $0.011$ & $0.005$ \\
& C (150)      & $1.000$ & $0.006$ & $0.002$ \\
& C (300)      & $1.000$ & $0.006$ & $0.002$ \\
\midrule
\multirow{3}{*}{Endo Helix (\(\theta_{\mathrm{endo}}\))} 
& Endo (60°)   & $0.998$ & $0.008$ & $0.003$ \\
& Endo (65°)   & $0.999$ & $0.007$ & $0.003$ \\
& Endo (70°)   & $0.999$ & $0.008$ & $0.003$ \\
\midrule
\multirow{3}{*}{Epi Helix (\(\theta_{\mathrm{epi}}\))} 
& Epi (60°)    & $1.000$ & $0.006$ & $0.002$ \\
& Epi (65°)    & $1.000$ & $0.006$ & $0.002$ \\
& Epi (70°)    & $0.999$ & $0.008$ & $0.003$ \\
\bottomrule
\end{tabular}
\end{table}

\subsection{Generalization to Unseen Global Physiological Values}

Table~\ref{tab:lovo_results} reported the results of LOVO validation across global physiological parameters, including pressure \(P\), stiffness \(C\), and fiber orientations \(\theta_{\mathrm{endo}}, \theta_{\mathrm{epi}}\). Despite the exclusion of specific values during training, \textit{HeartUnloadNet} consistently achieved high accuracy on the held-out cases, with DSC exceeding 0.998 and HD below 0.01\,cm in most settings. The only drop appeared at \(C=50\) (DSC: 0.942), suggesting a mild sensitivity to low-stiffness outliers and the importance of having sufficient training cases at outlying stiffnesses. Overall, the results demonstrated strong interpolation capability within the physiological range.

\subsection{Qualitative Visualization of Predictions}

\begin{figure}[htbp]
	\centering
	\includegraphics[width=\textwidth]{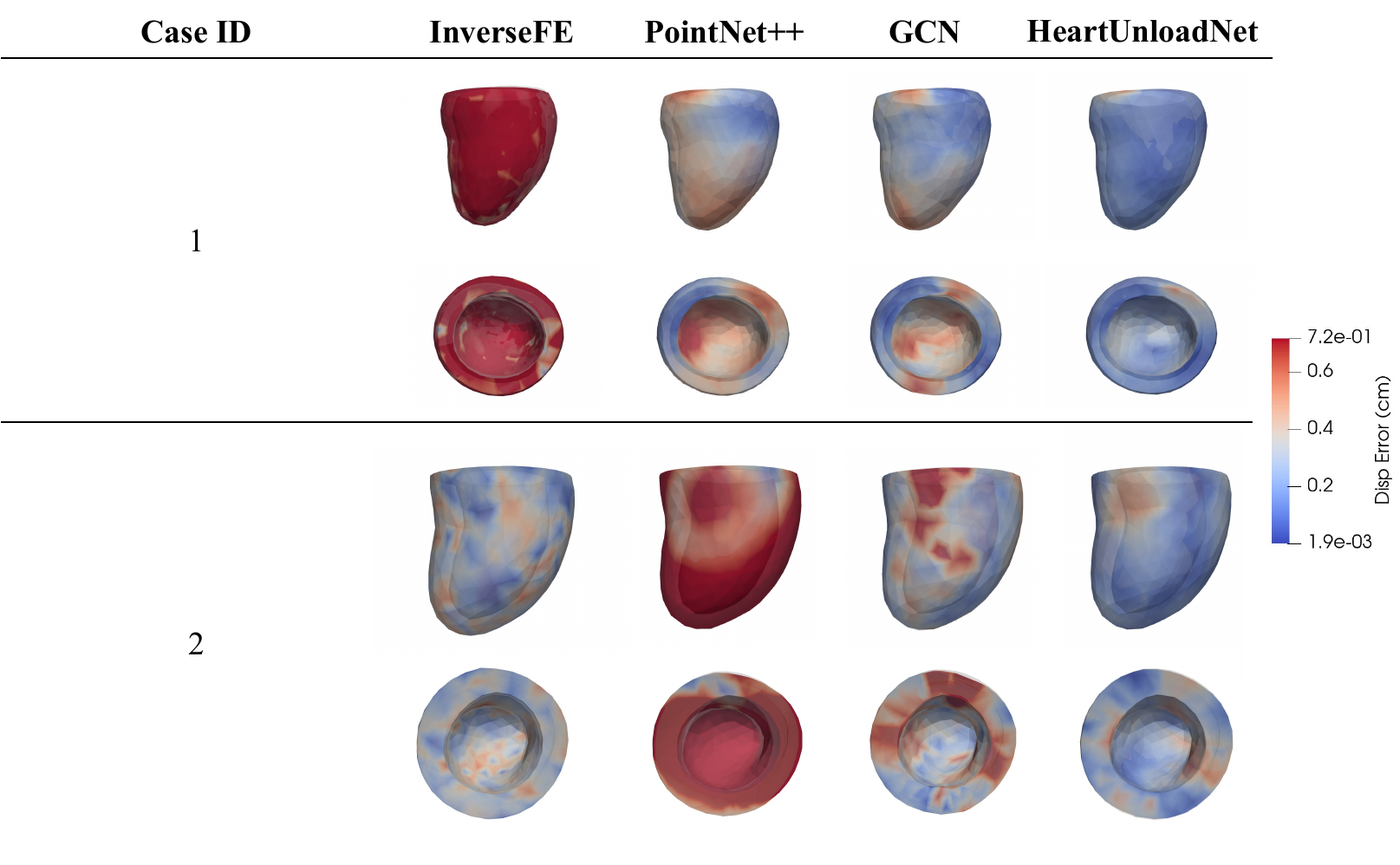}
        \caption{Qualitative heatmap visualization of vertex-wise prediction errors on two representative test cases. Each row shows a different method (Inverse FE, PointNet++, GCN and \textit{HeartUnloadNet}), and each column corresponds to one test case. Warmer colors indicate higher local displacement errors.}
	\label{FIG:vis}
\end{figure}

Figure~\ref{FIG:vis} presented heatmaps of vertex-wise displacement errors for two representative test cases from the test set. Given that the overall prediction error was on the sub-centimeter scale, direct visual comparison of mesh displacements was not intuitive. Therefore, we visualized the error distribution using colormaps rendered on the ED mesh to highlight subtle discrepancies. Warmer colors indicated larger local deviations, and all methods were compared under a unified scale and view orientation.\\

In Case 1, the inverse FE method exhibited globally elevated errors (reddish color), indicating a general mismatch between the predicted and true unloaded shapes. This was particularly visible from the basal view, where the inverse FE result showed notable shape distortion compared to the ground truth. For GCN and PointNet++, the errors were primarily concentrated around the apex and basal regions. In Case 2, PointNet++ performed poorly, with visibly misaligned predictions near the apex. GCN also showed moderate inconsistencies along the myocardial wall. In contrast, both inverse FE and \textit{HeartUnloadNet} performed better in Case 2, though \textit{HeartUnloadNet} maintained significantly lower errors (mostly under 0.1\,cm) in both cases, demonstrating clear superiority in both spatial precision and robustness.

\subsection{Advantages over PCA-Based and Inverse FE Approaches}
Table~\ref{tab:baseline-comparison} showed that \textit{HeartUnloadNet} reduced inference time by over $10^5{\times}$ while improving accuracy compared to the traditional inverse FE solver based on the Finsberg inverse step method \cite{finsberg2018efficient}. Although inverse FE can be accelerated using reduced-order models such as PCA-based statistical shape models~\cite{wang2020efficient}, these approaches remain limited in capturing complex unloading deformations.\\

To enable a fair comparison, we constructed a PCA model using displacement fields (i.e., node-wise differences between ED and unloaded meshes) across the training set. For each test case, we predicted the deformation by regressing ED-derived PCA coefficients and reconstructing the corresponding displacement field. The PCA-based predictions resulted in a HD of $1.02 \pm 0.07$\,cm. In contrast, \textit{HeartUnloadNet} achieved significantly lower errors, with a HD of $0.083 \pm 0.028$\,cm.\\

\begin{figure}
	\centering
	\includegraphics[width=0.7\textwidth]{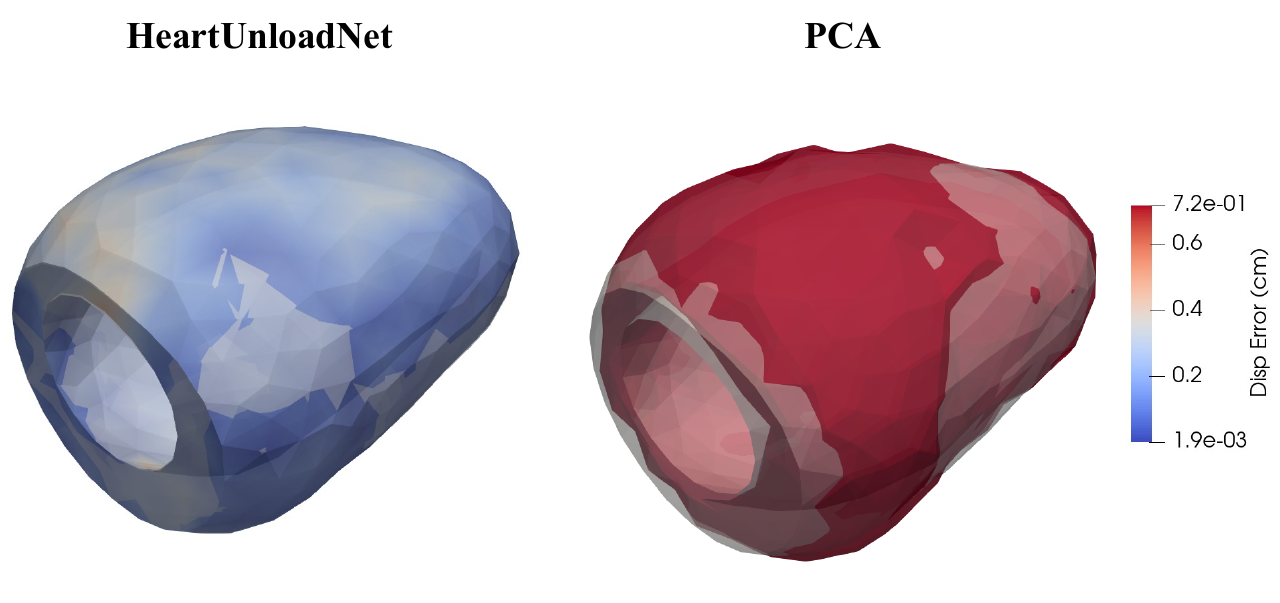}
        \caption{
        Qualitative heatmap visualization comparing vertex-wise prediction errors of PCA and \textit{HeartUnloadNet} for a representative test case. Compared to the ground truth, PCA exhibits a HD of $1.02 \pm 0.07$\,cm, whereas \textit{HeartUnloadNet} achieves $0.083 \pm 0.028$\,cm.
        }
        \label{FIG:disp_comparison}
\end{figure}

Figure~\ref{FIG:disp_comparison} visualized the deformation heatmaps predicted by PCA and \textit{HeartUnloadNet}. Ground truth vectors were omitted for clarity, as the predictions from \textit{HeartUnloadNet} were nearly indistinguishable from them. In contrast, PCA predictions exhibited clear discrepancies, highlighting its limitations in capturing subtle, spatially distributed deformations using a low-dimensional linear basis.\\

These observations highlighted the inherent limitations of PCA in modeling nonlinear tissue mechanics, where complex deformations could not be adequately represented by a small number of dominant modes. By learning a direct nonlinear mapping from ED meshes and physiological parameters to displacement fields, \textit{HeartUnloadNet} achieved not only higher accuracy and faster inference, but also better preservation of local biomechanical fidelity.

\section{Discussion}
\subsection{Novelty and Contribution}
The experimental results presented above validated not only the accuracy and efficiency of \textit{HeartUnloadNet}, but also demonstrated its potential to fundamentally shift the approach to unloaded geometry estimation. Unlike traditional inverse FE pipelines, which relied on iterative solvers, were sensitive to initialization and boundary condition settings, and exhibited slow convergence, our deep learning framework predicted the unloaded state through a single forward pass. \textit{HeartUnloadNet} consistently outperformed both Finsberg et al.'s inverse FE~\cite{finsberg2018efficient} and PCA-based reduced-order approaches~\cite{wang2020efficient} across all geometric metrics, and reduced inference time by over five orders of magnitude. These findings suggested that deep learning, when properly combined with mesh topology and physiological priors, was not merely a surrogate for inverse FE, but instead offered a robust and generalizable alternative for solving biomechanical inverse problems.

\subsection{Implications}
From a biomechanics perspective, accurately estimating the unloaded cardiac geometry was fundamentally important, as it served as the reference configuration in FE simulations of cardiac function. In applications such as heart disease analysis or interventional planning, all deformations, stresses, and energy quantities were derived from this baseline. Errors in the unloaded state propagated through downstream simulations, potentially distorting disease characterization, stress quantification, and virtual treatment assessments. A reliable unloaded geometry was therefore essential not only for passive mechanical analysis, but also for patient-specific inverse modeling and therapy optimization.\\

From a clinical perspective, fast and accurate estimation of the unloaded geometry could also support real-time, image-based cardiac analysis that may be developed in future. \textit{HeartUnloadNet} delivered near-instantaneous predictions with high geometric fidelity, making it well suited for integration into real-time diagnostic pipelines. As cardiac function assessment increasingly shifts toward image-driven, data-centric workflows, efficient estimation of the reference unloaded state will be a critical enabler for real-time stress or contractility evaluation.

\subsection{Architectural Strengths and Outlook}
A key advantage of our proposed framework lay in its mesh-agnostic design, which eliminated the need for a fixed nodal configuration or template mesh. Unlike many prior approaches that required point correspondences or mesh registration to enable learning, our method directly operated on arbitrary tetrahedral meshes, leveraging the inherent topology through GAT-based message passing. This flexibility allowed the model to accommodate diverse cardiac geometries, making it readily applicable to patient-specific data without labor-intensive preprocessing.\\

In addition, our architectural choices, including the use of graph attention, co-attentive global fusion, and cycle-consistent supervision, collectively contributed to the model’s superior prediction accuracy across all baselines. Beyond demonstrating performance, our study offered a comprehensive architectural evaluation, ablation and component-replacement experiments. Our results can provide insight into which architectural strategies were most effective for future for future model design in data-driven biomechanics.\\

We believe that \textit{HeartUnloadNet} offers a promising foundation for future development of end-to-end cardiac mechanics predictors. Although the current work focused on diastolic unloading, our GAT-based architecture can easily be extended to model dynamics during systole or across the cardiac cycles. Such future work might benefit further from incorporating physics-informed constraints, or residual formulations to jointly learn loading and unloading trajectories in a physiologically accurate, interpretable manner.

\subsection{Limitations}
\textit{HeartUnloadNet} demonstrated strong performance on data representing healthy anatomical variations. However, its generalizability to pathological cases remained to be further evaluated, and future work will investigate its robustness across a broader range of disease conditions.\\

\section{Conclusion}

In this paper, we introduced a data-driven framework for estimating unloaded LV geometry without relying on traditional physics-based optimization. By leveraging graph attention mechanisms and incorporating biophysical priors (pressure, stiffness, and fiber orientation), our model directly predicted deformation from ED meshes with high geometric fidelity. It eliminated the need for iterative FE solvers or reduced-order assumptions. On a held-out test set of 6,810 synthetic LV shapes, the network achieved a DSC of $0.986 \pm 0.023$ and a HD of $0.083 \pm 0.028$\,cm, with an inference time of just 0.023 seconds per case, surpassing inverse FE in both accuracy and efficiency.\\

\textit{HeartUnloadNet} also maintained strong performance under weak supervision, achieving over 97\% DSC with only 200 labeled samples. Its topology-agnostic design supported generalization across varying anatomical and material conditions, highlighting its potential for scalable application in both research and clinical settings. As the first learning-based method to infer unloaded cardiac geometry using explicit physiological priors, this work established a foundation for real-time, data-driven cardiac mechanics. Future extensions will integrate active contraction and systolic loading to enable full-cycle modeling.

\bibliographystyle{cas-model2-names}

\bibliography{cas-refs}

\end{document}